# Context based Roman-Urdu to Urdu Script Transliteration System

H Muhammad Shakeel[1], Rashid Khan [2,] and Muhammad waheed[3,]

[1]Comsats University of Information Technology Islamabad, Pakistan (Abbottabad Campus)
[2]University of Science and Technology of China (USTC), Hefei, People Republic of China
[3]Near East University Cyprus KKTC.

E-mail:  , hm_shakeel@gmail.com , rashidkhan@mail.ustc.edu.cn , muhammadwaheed2015@gmail.com

*Abstract*—Now a day computer is necessary for human being and it is very useful in many fields like search engine, text processing, short messaging services, voice chatting and text recognition. Since last many years there are many tools and techniques that have been developed to support the writing of language script. Most of the Asian languages like Arabic, Urdu, Persian, Chains and Korean are written in Roman alphabets. Roman alphabets are the most commonly used for transliteration of languages, which have non-Latin scripts. For writing Urdu characters as an input, there are many layouts which are already exist. Mostly Urdu speaker prefer to use Roman-Urdu for different applications, because mostly user is not familiar with Urdu language keyboard. The objective of this work is to improve the context base transliteration of Roman-Urdu to Urdu script. In this paper, we propose an algorithm which effectively solve the transliteration issues. The algorithm work like, convert the encoding roman words into the words in the standard Urdu script and match it with the lexicon. If match found, then display the word in the text editor. The highest frequency words are displayed if more than one match found in the lexicon. Display the first encoded and converted instance and set it to the default if there is not a single instance of the match is found and then adjust the given ambiguous word to their desire location according to their context. The outcome of this algorithm proved the efficiency and significance as compare to other models and algorithms which work for transliteration of Raman-Urdu to Urdu on context.

## I. INTRODUCTION

NLP stand for Natural Language processing. It is the Branch of computer science that deals with enhancement of the computers capability so that computer can easily understand the natural languages and it became easy for the human being to communicate with the computer system by using their natural languages. The capability of natural language program is to understand communication from user. Natural language processing is one of an application of artificial intelligence. Urdu is a burning area nowadays, which leads the ways to form emerging the innovative field of NLP. Urdu has the Indo-Aryan origin and is basically an Indo-European language. It is spoken widely across whole sub-continent Asia, but it is the mother language of Pakistan and also spoken in India while in other countries it is spoken as the secondary language. Urdu script is derived from Arabic and Pertain script hence it is aligned from left to right and shapes of words are too similar to the Arabic and Pertain words. For interface designing a natural language is used always, due to which many of the European languages are got much developed for interface designing. This created an inspiration for Urdu and other such languages. But still no research work like other languages is being done for Urdu. Our research paper is oriented around transliteration and context of ambiguous words. Transliteration is the technique for replacing words of source language into an equivalent target language. Particularly transliteration is a very helpful tool that is used by most of the applications (i.e information extraction (IE), search engine, text processing, short messaging services and cross language information retrieval (CLIR) etc.) [1]. Transliteration is also used for the languages to translate named entities. When we are Translating a word from its origin language to a foreign language by using transliteration then we can say that it is forward transliteration. While when we are translating a word from foreign language to back to the origin language then we can say that it is backward transliteration. As we are replacing word of one language to another language by using transliteration so the exact translation is not possible, for example in Urdu language there are some that does not have an equivalent word in English i.e muhajir (the word is used for the people that migrate from other countries ) so in English there  is not an equivalent word that is inserted for muhajir.There are some issues occur when we are translating using transliteration that is when absolute equivalence rather than relative equivalence is required. Sometime a situation occurs then we can't compare in loss of information then obviously not only the translation but the overall communication is not possible.

### A. Urdu Language

Urdu the national language of Asian nation and belongs to the family of Indo-European languages. The word Urdu    is essentially derived from a Turkish word ORDU or ORDA which suggests ARMY. The morphology of Urdu language is extremely troublesome owing to the inheritance completely different of various words from different languages round the world like Sanskrit, Punjabi, Arabic, Turkish, and English etc. The vocabulary of Urdu is developed by mistreatment Persian and Sanskrit and conjointly there was influence of Turkish and Arabic as well owing to characteristics like word borrowing and morphology, Urdu is additionally called the language of poets. Thus, Urdu is classified because the right to left language that need correct usage of font and Unicode cryptography for process it. Not like Asian nation, Urdu is additionally spoken in numerous elements of Bharat, China, South Korea USA, Britain and therefore the Middle East. Urdu is stratified as twentieth most generally oral communication within the World several of linguistic counts Urdu and Hindi because the same language owing to the massive similarity within the descriptive linguistics employed in each languages. The Weber's article on high languages states Urdu/Hindi as fourth most generally oral communication round the world consistent with him, 4.7 of the globe population speaks Hindi/Urdu. The word Urdu derive from Turkish language, which mean horde i.e. Lashkar. There are eleven million peoples in Asia who speak Urdu language. It is much closed to Hindi language among all other languages. Urdu and Hindi each have originated from the accent of metropolis region and apart from the minute details these languages share their morphology. Since there are many words of Sanskrit which are used in Hindi, similarly vide range of Arabic and Persian vocabulary that are used in Urdu language [2]. English, Turkish and Portuguese vocabulary are also use in Urdu language. There square measure quite ranges of words that have found an area in Urdu Language, typically through the Persian Language, have otherwise nuanced connotations and usages.

### B. What is Transliteration

The procedure of converting or transcribing phrase from one script to any other script such that focus on script phrases are phonetically equal to the source script words. Many of the Urdu speakers lack proficiencies in English, thus creating a barrier for such person in this advance world to communicate through the emails and SMS service. So this transliteration software for considering the peoples of South Asian like Pakistan, India and Bangladesh etc. As it will convert the text written in the roman Urdu to legitimate Urdu text at the sender end before sending so that receiver can get the text written in the Urdu script. The roman script for writing Urdu is not a standard or official script, it is adopted widely due to the following major reasons: The default interface of the computer and the handheld device such as android phones and tablets etc. is in English. The wide influence of English language, especially on the educated community. English is the official language of Pakistan and an international language. Writing in standard Urdu script needs installation of separate of Urdu support. People may found burden to type and look for alphabets to which are not use to of typing. In this third world, technology is developing rapidly and computation power is increasing drastically which helps in solving many multilingual bilingual problems in the language. During informal communication and chatting, mostly people not know to install the Urdu support software so transliteration is a good technique in such case [3]. Another approach considered to be adopted for English to Urdu Transliteration is rule base orthography mapping, but unfortunately, it was soon revealed that it cannot work as there isn't any one-to- one mapping between English alphabets and sounds as for 26 alphabets there are 44 sounds [4]. Writing any language in its standard format is the basic necessity of the native speakers. Over the number of years, the tools have been developed. IME editors here focus on getting the value through the keyboards, it also helps in transliteration process for converting the roman text to Urdu text. Transliteration is a sub-field of computational linguistics, and its language processing requirements make the nature of the task language-specific. Although many studies introduce statistical methods as a general-purpose solution both for translation and transliteration, specific knowledge of the languages under consideration is still beneficial. Generally, a transliteration system inputs a word in a source language and outputs a word in a target language which is pronounced the same as the source word. This process is also called phonetical translation. A transliteration system performs the transformation using a transliteration model, created specifically for the source and target languages. Since the existence of such a model is crucial, the transliteration that a system provides becomes specific to the language-pairs on which it is trained.

### C. Issue in Urdu Transliteration

Roman script for Urdu is widely used by the Urdu speakers but has no standard format of spellings, many of the words a r e randomly spelled, so a single word can have more than one spelling as every person use different spellings due to absence of any standard and grammar rule, they just make a cluster of alphabets which can produce the same sound as the original Urdu word sounds. Another major problem is that the total number of sounds and alphabets in English and Urdu are different so, there is no clear cut mapping between the English and Urdu alphabets and sounds. No significant work has done for automatic transliteration of the Roman script to Urdu script for Urdu text. This translation achieved through character mapping and word mapping, which has been discussed earlier.

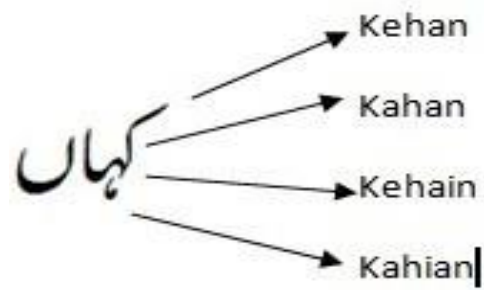

Fig. 1. This shows many Roman Urdu against one Urdu word

### D. Applications of Transliteration

Transliteration is widely used for the transliteration of one natural languages script to another language. The Transliteration experts have done a lot of research in this field, but most of them are done in English and chains languages. Most of t h e transliteration is use for internet chatting, mobile chatting and search engine optimization. Related work

## II. RELATED WORK

Many of the Asian languages like Chinese, Arabic, Hindi, Persian, Urdu, and Pashto are written in roman script especially during the mobile chat or internet chat. Urdu written in roman script is called Roman-Urdu. The reason for using the roman script for Urdu is that as all the devices have default user interface in English. So users are familiar with the keyboard layout of Urdu, so they avoid using Urdu script despite the availability of the number of keyboard layouts of Urdu. Transliteration is the solution to the problem, which is the transformation of the script from one language to another language in the phonetically equivalent script in another language. There exist a number of transliteration schemes, but like all languages with non-Latin script these roman letters are used for mapping the script. Though small work has done till for Roman script to Urdu script transliteration but still different models has been purposed for such transliteration. In these existing models character by character mapping is done but both the scripts of Urdu i-e Roman Urdu script and the Standard script does not have the same number of letter, so direct mapping is not good or impossible in some case, hence symbols are used for mapping instead of directly mapping letters. N-Gram is an algorithm which predicts the next words using probabilities. This algorithm examines n-1 word and then give the prediction of the next word, which help to use the next word to context base [5]. Urdu transliteration is extremely specialized one and is single schema system but still not the better approach. There are many similarities in needs for the transliteration system in Urdu and Persian language, for which a better translation system exist. Actually, that transliteration system is based on statistical modeling approach, thus eliminating the need for symbols, and handling the diversity of spellings. Some work done for Urdu script transliteration includes, Tafseer Ahmed and Bushra Jawaid research work, after a deep literature review and critical analysis of the existing Transliteration system for Hindi to Urdu scripts they suggested required enhancements and improvement in the translation system [6]. They mainly focused the character by character mapping of the letters. Tafseer Ahmed also developed a world list for Raman Urdu Script to standard Urdu script Transliteration with better results [7]. Christian Boitet and Abbas Malik used a parallel word list for analyzing the statistical Machine transliteration for Hindi to Urdu transliteration purpose [8]. MANOJ Ketal another researcher used character sequence modeling (CSM) for exploiting the performance of the transliteration results. He proved that proper joining of the monolingual resources will result in far better results. Usman Afzal also had developed a translation model for roman Urdu to standard Urdu known as Cross-Script Trie Generation model, but this model is dictionary based which enforces certain limitations like, a word which is not present in the dictionary cannot be transliterated. Sill context based transliteration is neglected are in Roman-Urdu Script to Standard Urdu script transliteration. Generative and descriptive methods are two most widely used methods and a lot of research work had been done for improving their performance [9]. There was a huge amount of research targeted at the assignment of transliteration with each discriminative and generative techniques accomplishing applicable usual performance. Ravi and Knight proved the quit end result with the generative version for Eastern-English fashions on unsupervised data. Graehl and Knight proved results of back-transliterating from Japanese to English far better with the generative model while using a weighted FST. Traditional source-channel SMT model is another model used by the Khudanpur and Virga for translation of sounds from one language to another and proved the results for Chinese to English transliteration. Phoneme-based transliteration model (PTM) consists of two phase-in first phase source graphene are transformed source phoneme and in the second phase source phoneme are transformed to target grapheme. Graehl and Knight had modeled Japanese-to-English transliteration using a similar approach. They used weighted finite-state transducers (WFSTs). They first transformed first from Romani-to-phoneme, after which within the 2nd segment from phoneme-to-English. Right here they combined several parameters and purposed a similar model for proposed an English-to-Chinese transliteration which turned into primarily based totally on Chinese language phonemes, English grapheme-to-phoneme conversion, mapping regulations among English and Chinese language, Chinese syllable-based totally, move-lingual phonological and individual-based totally definitely language models. Junget provided each other version for English-to-Korean transliteration which used extended Markov hf [10]. Wherein English phrase is first transformed into English pronunciation with the assist of a pronunciation dictionary. After which English phonemes segmented into the chunk of English phonemes. Now predefined handcraft recommendations are used right here with the assist of which every phoneme corresponds to as a minimum one Korean grapheme. On the surrender prolonged Markov window are used for final transformation of every bite of English phoneme into Korean graphemes. All the natural languages have its own set of alphabets and sounds, it's not necessary that every sound in one language should correspond to a sound in the other language. Hence, these missing sounds are replaced by two letters known as digraphs or by three letters tri graphs. The location of system transliteration has been substantially studied at the same time as focusing one-of-a-kind languages which advanced a number of techniques, Grapheme approach and phoneme approach are two extensively used processes. Every other transliteration gadget with the aid of Haque et.al for Hindi- to- English transliteration used PB-SMT model that is developed by using the phrase-based totally statistical approach, this version consequence in the development of 43.44 % for general data set at the same time as approximately 26.42% for large datasets Jia, Zhu, and Yu used Grapheme-primarily based device Transliteration for developing their Noisy Channel model. The model changed into implemented through the usage of Moses (a translation device). Each ahead and backward transliteration between English and Chinese language is carried out in this system. Chinnakotla, Damani, Satoskar used character sequence modeling for the aid-scarce languages. This model reported the efficiency of 70.7 % Hindi to English, 67 % English to Hindi and 48.0 % for Persian to English transliteration [11]. This method first at the source side uses CSM perceive the word beginning, then for producing transliteration candidates it manually generated no probabilistic character mapping rule base finally once more makes use of the CSM for ranking the generated candidates at the target aspect [12]. Chinnakotla and Damani had additionally developed Transliteration structures for English to Kannada, English to Hindi and English to Tamil.

### III. Proposed Methodology

In the proposed solution we have developed an algorithm which is used for transliteration of Roman-Urdu to Urdu script. The main objective of this research is to develop an algorithm that can improve the result for the transliteration of Roman-Urdu. The existing system does not work with context base script; therefore, the result of the given work is not satisfactory. Due to very little research done in the field of NLP for the

Urdu language, we are unable to get a standard sized corpus for the Urdu language.

*A. Proposed algorithm*

The following algorithm will help for using context base sentence, as for each sentence we can have multiple multiple ambiguous words. This algorithm consists of different steps, which are:

1. First get input from user.
2. Read input and split it on the basis of sentence terminators.
3. Give sentence list to array.
4. Apply loop on each array item for transliteration from API.
5. Now check ambiguous word in each sentence.
6. If ambiguous word not exists in sentence then skip and check for other sentence.
7. If ambiguous word exist in sentence then its characteristics and match sentence words with its characteristics.
    - if match with characteristics then skip store count match and check for next word in group until last word.
    - check the count larger value (maximum matches ) where maximum matches replace it with ambiguous word.
8. Display result of translated array.

Fig. 2. an algorithm for Transliteration

There are sounds base ambiguous words in Roman Urdu , to improve the accuracy of transliteration we handle these words. The outcome of these words is given below.

## IV. EVALUATION

We will evaluate the accuracy and correctness of software. The criteria adopted for this purpose is evaluation metrics. This section also contains discussion about the experimental results. To check the efficiency of our algorithm, we have tested different paragraphs and we got the results which are plotted in this paper.

## V. CONCLUSION AND FUTURE WORK

This system is the full-fledged working system but still a number of features can be added to it for enhancing it functionalities and making its services more reliable and user-friendly. Adding dynamic lexicon is still our major concern. We firmly state that our system will full fill the user requirements in a user-friendly way and generates results with the higher degree of accuracy. These types of rules can be implemented for Arabic, Persian, and Hindi. Our system consists on context base by Using API. It cannot transliterate ambiguous words which is occur twice in sentence, which is limitation of our system. In roman English script sometimes we write two words without using space but in Urdu these words are two separate words in Urdu script. Some examples are (ap ko , jb k , milkr hints ya urdu a lekhnay hain) which are written either as jbk and mil kr or alternatively as ap ko, jb kand mil kr etc.

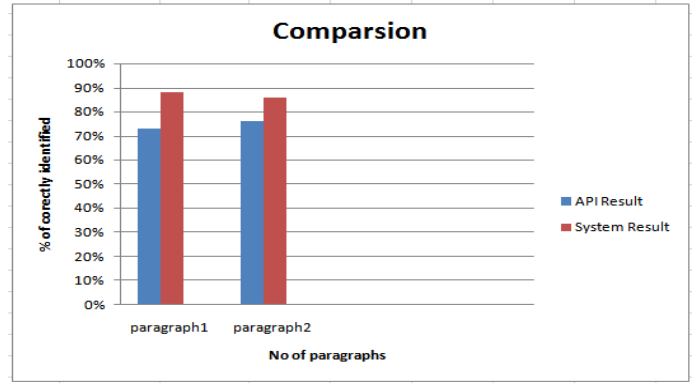

Fig. 3. Result of paragraph1 and paragraph2

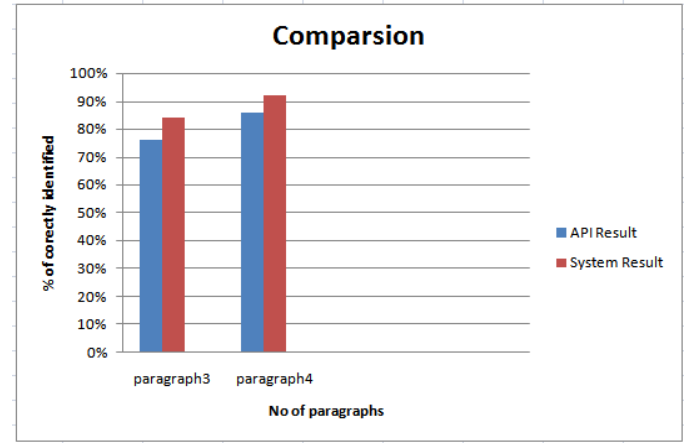

Fig. 4. Result of Paragraph 1

Future work:

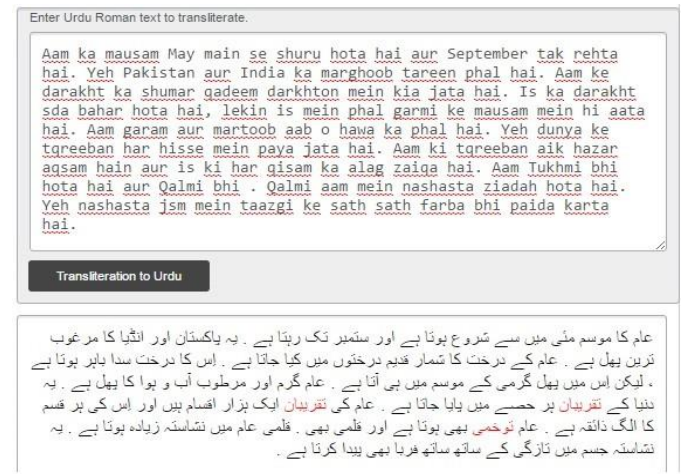

Fig. 5. Context base Roman Urdu output of Ijunun

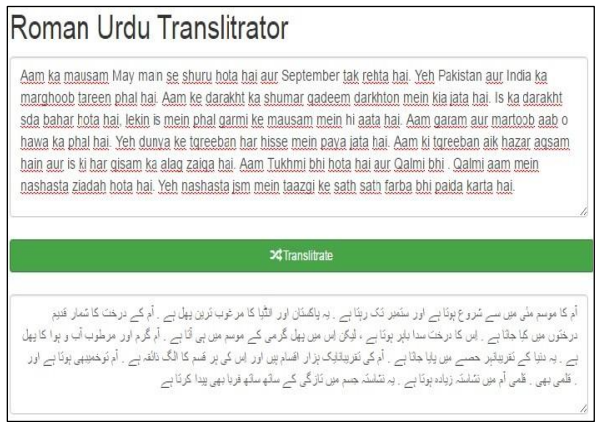

Fig. 6. Context base Roman Urdu output of System

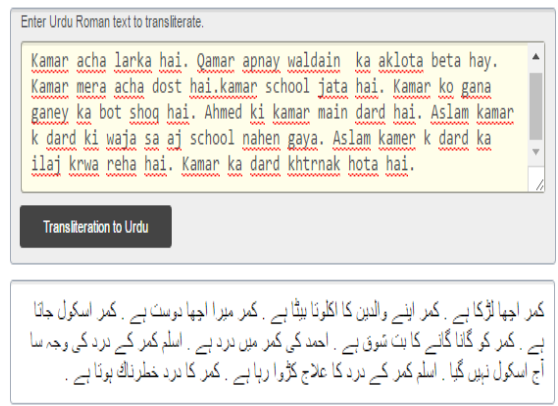

Fig. 8. Context base Roman Urdu output of IJunun

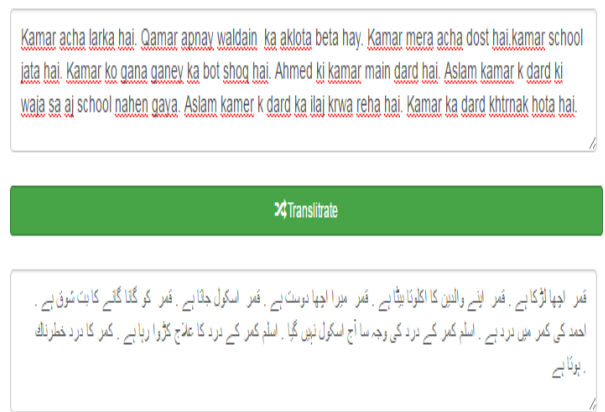

Fig. 7. Context base Roman Urdu output of System